\pdfoutput=1

\documentclass[11pt]{article}

\usepackage[preprint]{acl}

\usepackage{times}
\usepackage{latexsym}

\usepackage[T1]{fontenc}

\usepackage[utf8]{inputenc}

\usepackage{microtype}

\usepackage{inconsolata}

\usepackage{graphicx}

\usepackage{amsmath}
\usepackage{amssymb}
\usepackage{mathtools}
\usepackage{algorithm}
\usepackage{algorithmic}
\usepackage{subcaption}
\usepackage{booktabs}
\usepackage{bm}
\usepackage{multirow}
\usepackage{multicol}
\usepackage{colortbl}
\usepackage{arydshln}
\usepackage{stfloats}

\definecolor{lightgreen}{RGB}{200,255,200}

\def\vtheta{\bm{\theta}}
\def\vdelta{\bm{\delta}}
\def\vtau{\bm{\tau}}
\def\vgamma{\bm{\gamma}}

\title{\textsc{LoRE-Merging}: Exploring Low-Rank Estimation For Large Language Model Merging}

\author{Zehua Liu$^{1}$, Han Wu$^{1}$, Yuxuan Yao$^{2}$, Ruifeng She$^{1}$, Xiongwei Han$^{1}$ \\ \textbf{Tao Zhong}$^{1}$, \textbf{Mingxuan Yuan}$^{1}$\\
$^{1}$ Huawei Noah's Ark Lab\\
$^{2}$ Department of Computer Science, City University of Hong Kong\\
\texttt{liuzehua@connect.hku.hk}\\
\texttt{wu.han1@huawei.com}\\
}

\begin{document}
\maketitle
\begin{abstract}
While most current approaches rely on further training techniques, such as fine-tuning or reinforcement learning, to enhance model capacities, model merging stands out for its ability of improving models without requiring any additional training.
In this paper, we propose a unified framework for model merging based on low-rank estimation of task vectors without the need for access to the base model, named \textsc{LoRE-Merging}. Our approach is motivated by the observation that task vectors from fine-tuned models frequently exhibit a limited number of dominant singular values, making low-rank estimations less prone to interference. We implement the method by formulating the merging problem as an optimization problem. Extensive empirical experiments demonstrate the effectiveness of our framework in mitigating interference and preserving task-specific information, thereby advancing the state-of-the-art performance in model merging techniques.
\end{abstract}

\section{Introduction} \label{sec:intro}

Large Language Models (LLMs) have become ubiquitous in numerous real-world applications \citep{bommasani_2021_opportunities, zhuang_2020_comprehensive}. The utilization of LLMs typically involves fine-tuning them for specific tasks, a process that often yields superior performance compared to general-purpose LLMs. A rapidly emerging technique in this domain is model merging \citep{garipov_2018_loss, wortsman_2022_model,yu_2024_language}, which aims to create a single multi-task model by combining the weights of multiple task-specific models. This approach facilitates the construction of multi-task models by integrating knowledge from fine-tuned (FT) models without requiring additional training. 

Building on recent studies \citep{ilharco_2022_editing, yadav_2024_ties, yu_2024_language}, task vector-based merging approaches have demonstrated significant effectiveness, where task vectors are defined as the parameter differences between fine-tuned models and the base LLM.
Achieving optimal results in model merging often requires minimizing interference between task vectors associated with different tasks. To address this, existing approaches utilize modified task vectors instead of the original ones. For instance, \citet{yu_2024_language} applied random dropping with probability $p$ to obtain a sparse representation of task vectors, while \citet{yadav_2024_ties} retained only the top-$k$ elements of each task vector based on magnitude, setting the remaining elements to zero. These strategies aim to produce sparse estimations of task vectors, a common technique for mitigating interference.

Nevertheless, task vector-based model merging approaches remain constrained by two fundamental limitations. First, the computation of task vectors necessitates access to the base model parameters and demonstrates heightened sensitivity to parametric variations \citep{yu_2024_language}. As fine-tuning progress goes deeper, substantial parametric divergence emerges between the original base model and its fine-tuned counterpart, thereby greatly hindering them merging effectiveness \citep{yu2024extend}.
Second, empirical evidence from \citet{yadav_2024_ties} reveals that conflicting task vectors interactions could appear even when employing sparse estimation techniques. On the other hand, the sparsification process risks inadvertently eliminating essential task-specific features, thereby compromising the efficacy of the resultant merged model. These inherent constraints of sparse approximation methodologies underscore the necessity for developing alternative frameworks to estimate higher-fidelity low-rank task vector representations.


\begin{figure}[t!]
\centering
\includegraphics[width=0.23\textwidth]{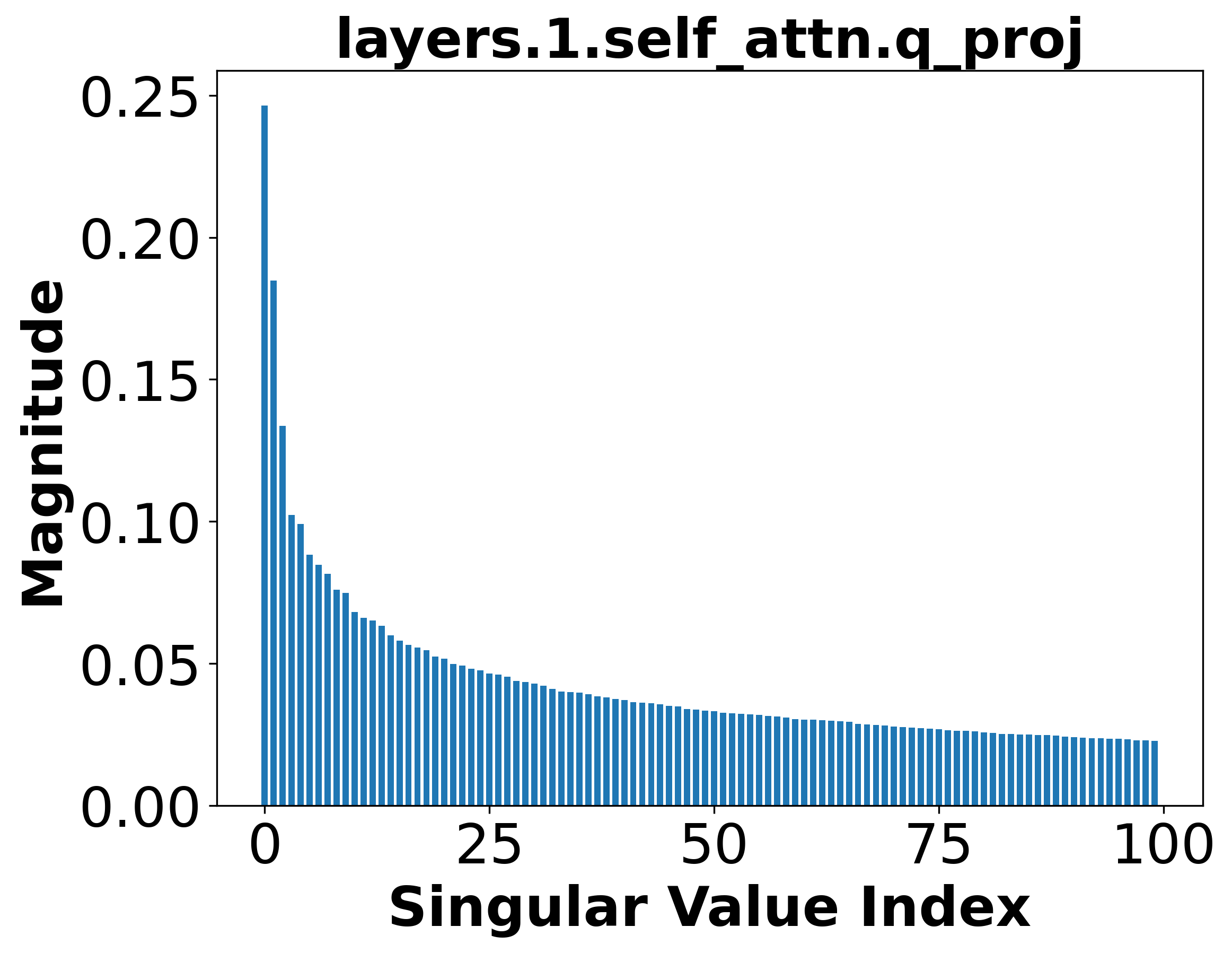}
\includegraphics[width=0.23\textwidth]{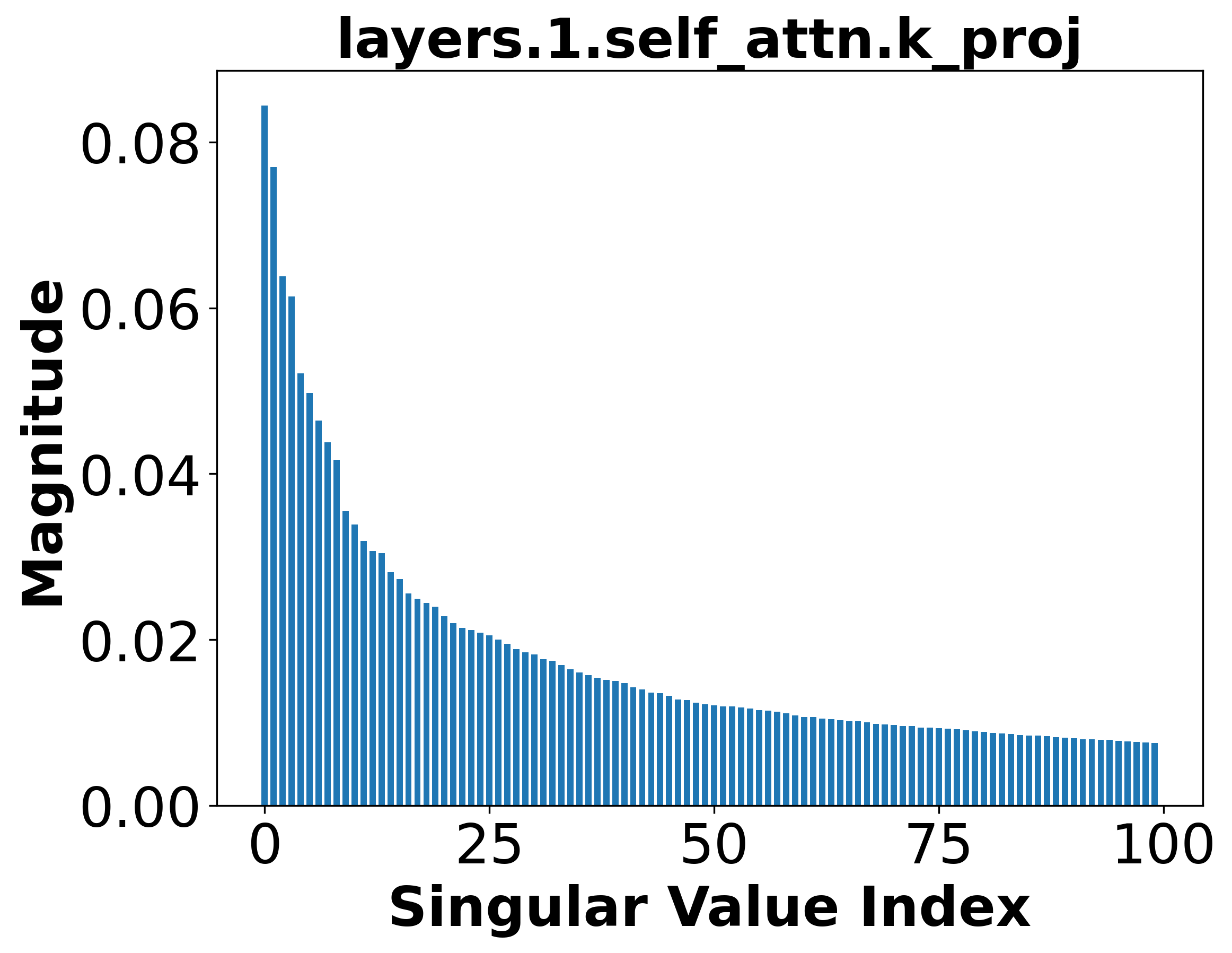} \\
\includegraphics[width=0.23\textwidth]{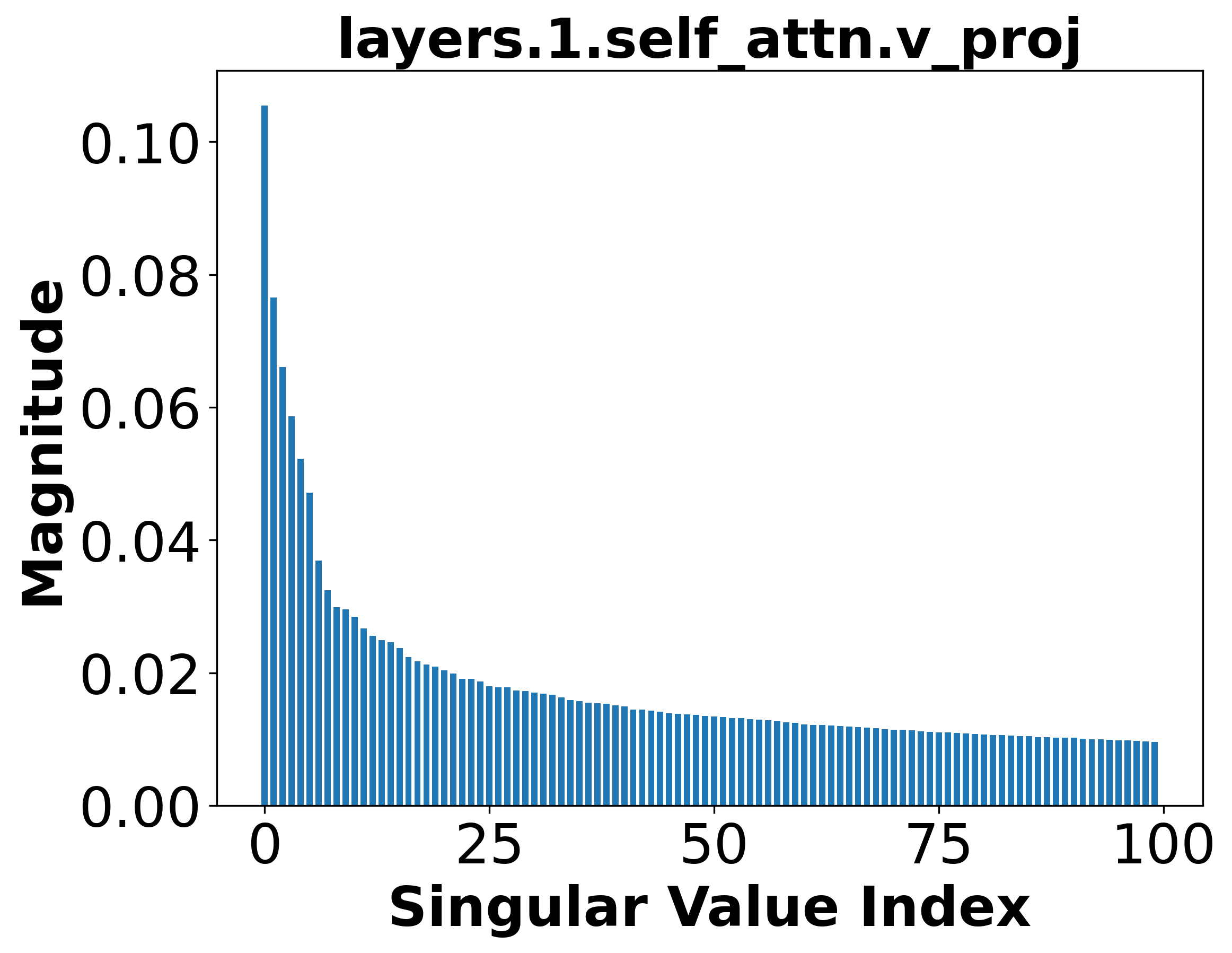} 
\includegraphics[width=0.23\textwidth]{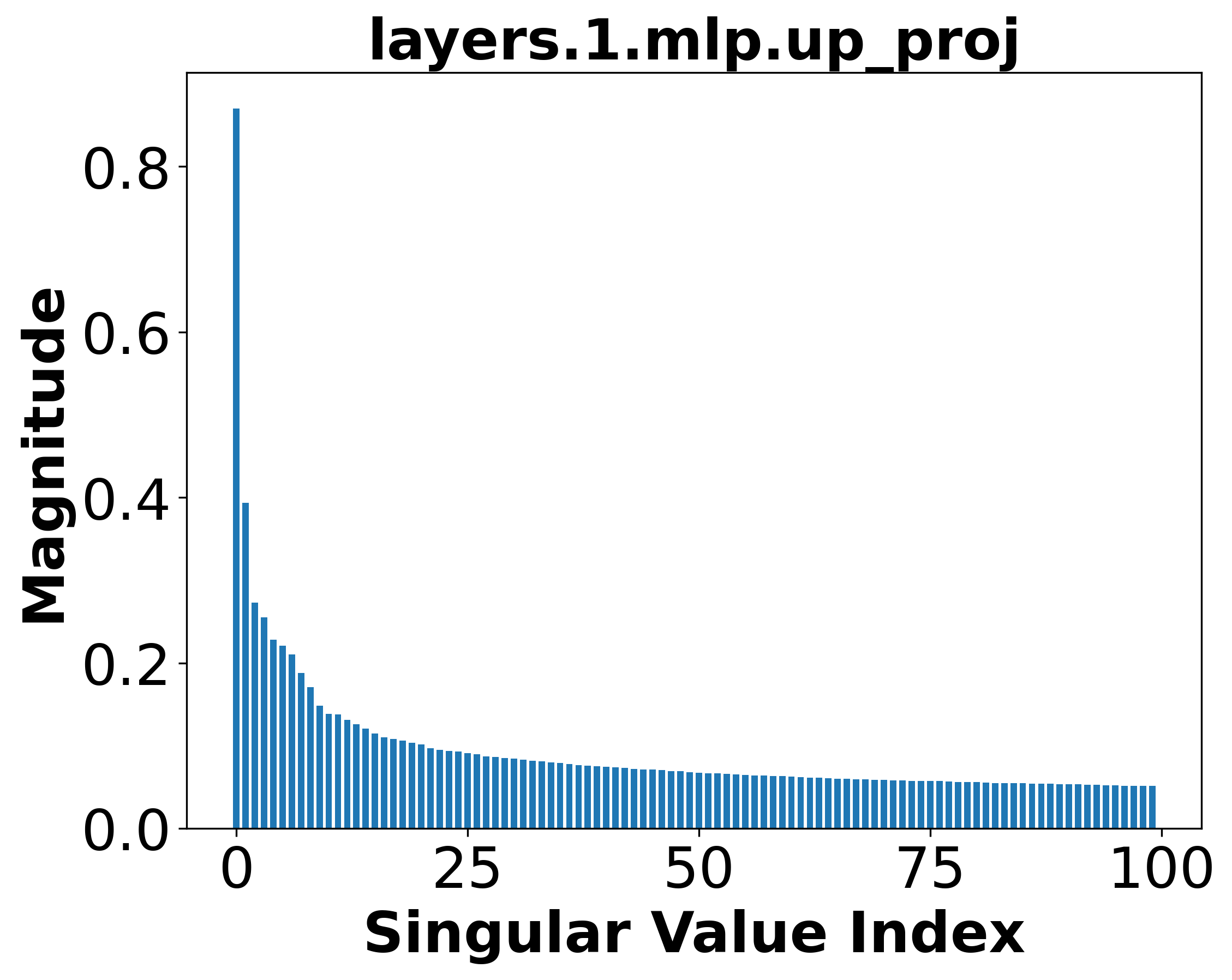} \\
\includegraphics[width=0.23\textwidth]{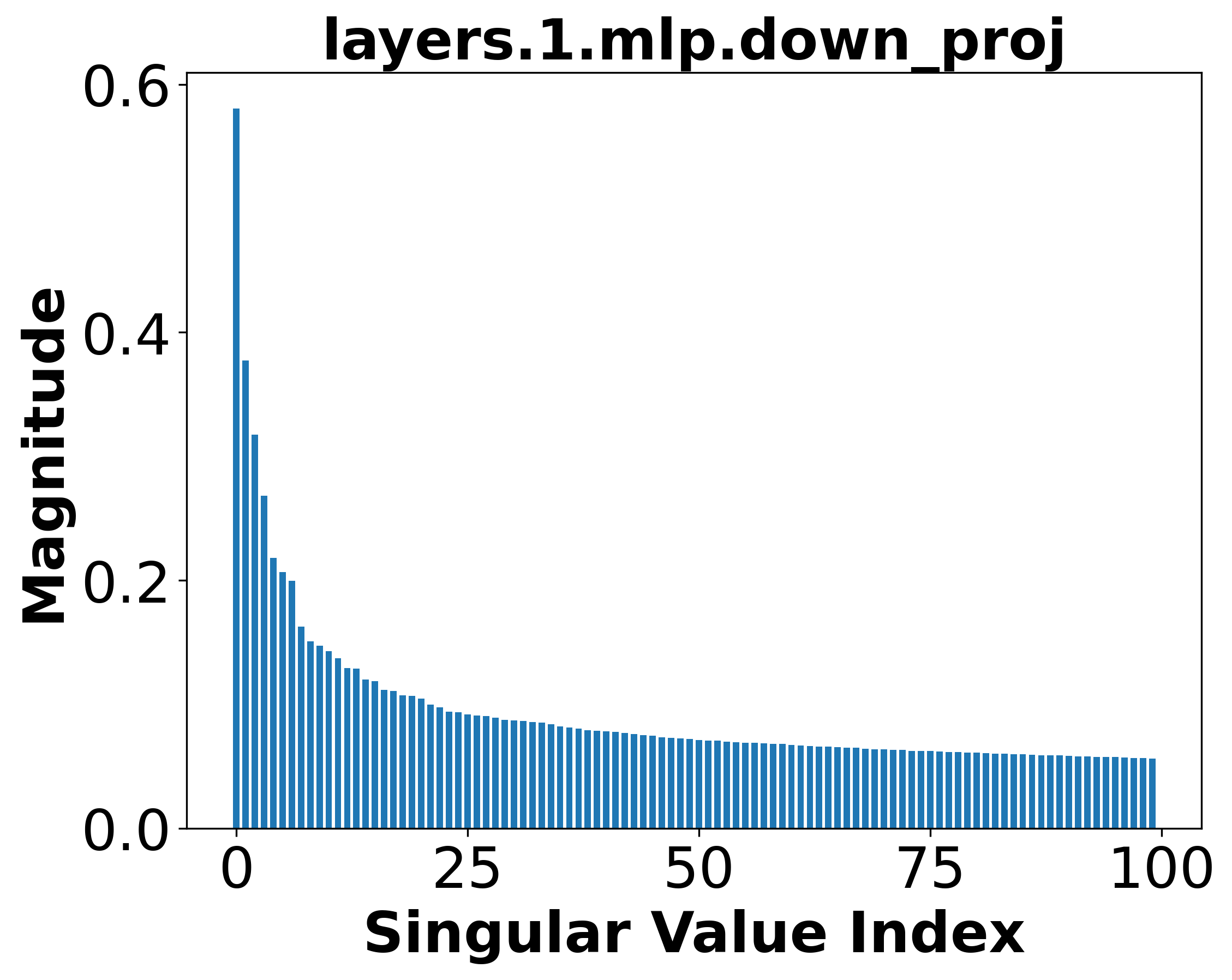} 
\includegraphics[width=0.23\textwidth]{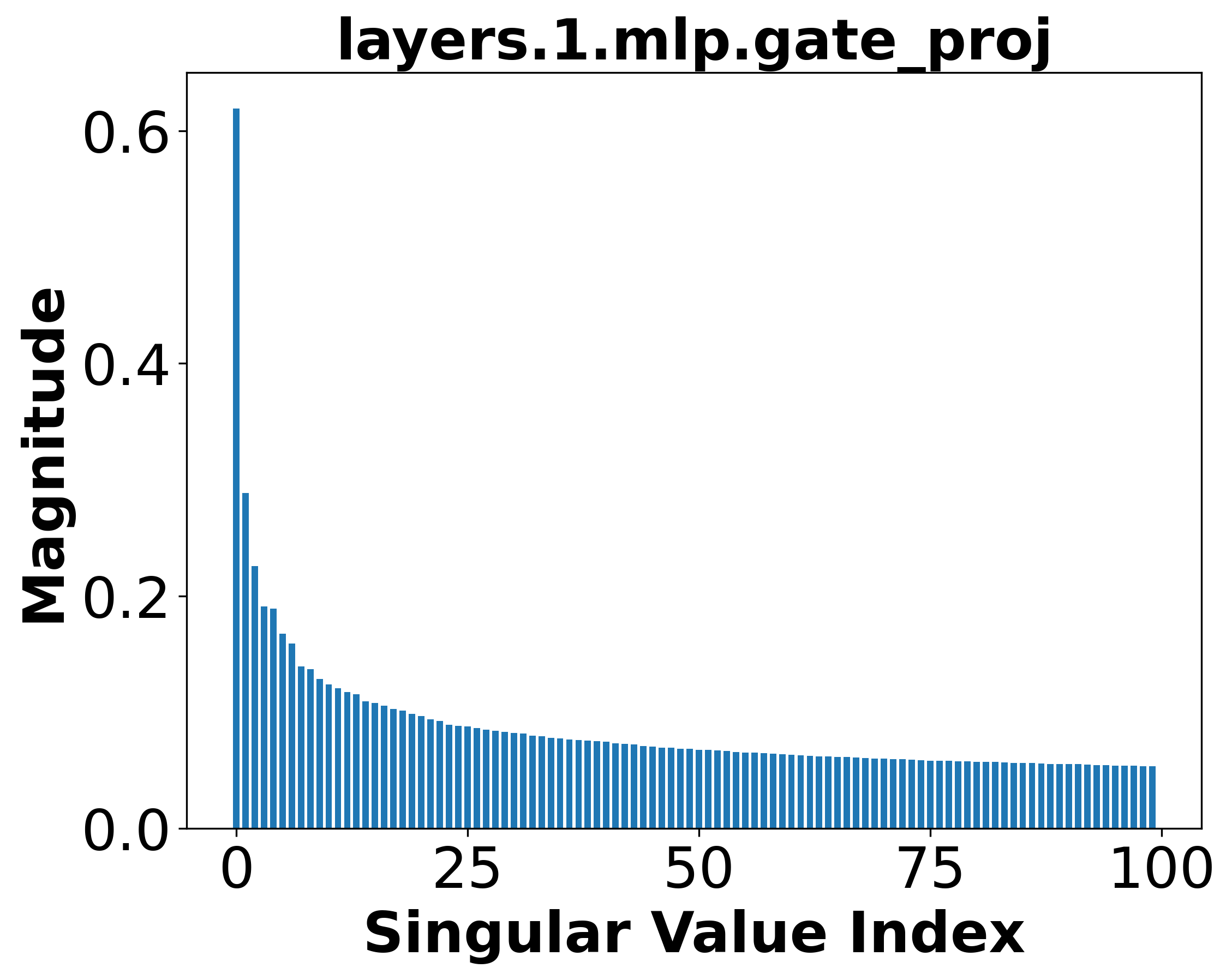}
\caption{Singular value distributions for the task vector in layer $1$. We show the top-100 singular values, out of 4096 within the full rank.}
\label{fig:1}
\vspace{-1em}
\end{figure}

To this end, we first empirically validate that task vectors exhibit a small number of dominant singular values, with the remaining singular values being significantly smaller in magnitude, as shown in Figure \ref{fig:1}. Additionally, the dimension of the intersection of the images of two matrices is bounded by the minimum of their ranks. Therefore, we propose \textsc{LoRE-Merging}, a unified framework for model merging based on \textbf{Lo}w-\textbf{R}ank \textbf{E}stimation of task vectors, which eliminates the need for access to the base model.
Specifically, given a set of FT models, we formulate the merging problem as an optimization problem whose goal is to simultaneously identify an approximate base model integrated with a set of low-rank task vectors. Together, these vectors collectively approximate the behavior of the FT models.
By leveraging low-rank estimations, task vectors become inherently less susceptible to interference, effectively addressing a fundamental challenge in model merging.
We conduct extensive experiments on optimization modeling problems and math word problems to confirm the effectiveness of our method.




\section{Related Works}

Merging fine-tuned models has been shown to offer several benefits, such as improving performance on a single target task \citep{gupta_2020_stochastic, choshen_2022_fusing, wortsman_2022_model}, enhancing out-of-domain generalization \citep{cha_2021_swad, arpit_2022_ensemble, ilharco_2022_editing, rame_2023_model}, creating multi-task models from different tasks \citep{jin_2022_dataless, li_2022_branch, yadav_2024_ties}, supporting continual learning \citep{yadav_2022_exclusive, yadav_2023_exploring}, and addressing other challenges \citep{don_2022_cold, li_2022_branch}.
Among these methods, task-vector-based merging approaches play an important role. Task Arithmetic \citep{ilharco_2022_editing} first introduced the concept of task vectors and shows that simple arithmetic operations can be performed to obtain the merged models. Building on this idea, methods like DARE \citep{yu_2024_language} and TIES-Merging \citep{yadav_2024_ties} adopt pruning-then-scaling techniques to merge task vectors, based on the assumption that not all parameters equally contribute to the final performance. However, these methods based on sparsity estimation consistently suffer from the interference among task vectors and require access to the base model, thus limiting their overall effectiveness.


\section{Methodology}
\subsection{Problem Setting}

We denotes $\mathcal{M}_i$ as the candidate models to be merged, where each $\mathcal{M}_i$ is parameterized by $\vtheta_i$. 
In this work, we focus on the homogeneous model merging \citep{wortsman_2022_model, ilharco_2022_editing, yadav_2024_ties}, suggesting that the base models share the same model architecture. Specifically, these models can be obtained from the training process, such as checkpoints, or fine-tuned from the same pre-trained model, referred to as task-specific models.
The primary objective of model merging is to construct a new model, $\mathcal{M}^*$, having better performance on the target single or multiple tasks.

\subsection{Implicit Low-Rank Estimation for Model Merging}

In this study, drawing upon methodologies similar to those presented by \citet{matena_2022_merging}, we investigate the model merging problem without presupposing specific characteristics of, or requiring access to, a base model.
This methodological decision is underpinned by several key rationales.
Firstly, in the context of checkpoint merging \citep{liu_2024_checkpoint}, a prevalent scenario involves access restricted solely to checkpoints saved during the training trajectory, before the finalization of a base model.
Consequently, in such instances, the assumption of a pre-defined base model is untenable.
Furthermore, as demonstrated by \citet{yu_2024_language, yu2024extend}, model pairs frequently exhibit limited mergeability, particularly when subjected to extensive fine-tuning or prolonged pre-training, which can induce substantial parametric shifts.
Under these circumstances, existing task-vector-based merging techniques often prove less effective due to significant representational divergence between an original base model and its fine-tuned counterpart.
To surmount this challenge, we introduce \textsc{LoRE-Merging}, an implicit low-rank estimation approach to model merging. This method leverages the inherent robustness of low-rank estimation against perturbations while obviating the requirement for base model access.

The core idea of \textsc{LoRE-Merging} is straightforward: instead of using the original base model, we first construct an approximate base model and subsequently integrate the task-specific vectors via a low-rank approximation technique.
Formally, denote the approximate base model as $\vtheta_0$ and the low-rank task vectors $\{ \vdelta_i \}_{i=1}^n$ where $n$ is the number of FT models, our objective is to minimize the discrepancy between each FT model and its corresponding integrated version derived from the constructed base model, expressed as $\vtheta_0 + \vdelta_i \approx \vtheta_i$.


To ensure the low-rank structure of $\vdelta$, we apply a nuclear norm penalty, as suggested in \citet{cai_2008_singular}. Then, we formulate the merging problem as the following optimization problem:
\begin{equation} \label{equ:2}
\min_{\vtheta_0, \vdelta_1, \dots, \vdelta_n} f := \sum_{i=1}^n \left( \| \vtheta_0 + \vdelta_i - \vtheta_i \|_F^2 + \mu \| \vdelta_i \|_*^2 \right),
\end{equation}
\noindent where $\| \cdot \|_*$ represents the nuclear norm, and $\mu > 0$ is a hyperparameter.
In Equation (\ref{equ:2}), the first term minimizes the difference between $\vtheta_0 + \vdelta_i$ and $\vtheta_i$, ensuring reconstruction accuracy. The second term acts as a penalty that encourages the task vectors $\vdelta_i$ to exhibit low-rank properties.

This problem is a standard multi-variable convex optimization problem. To solve it efficiently, we employ the coordinate descent method \citep{wright_2015_coordinate}.
Starting from an initial point $\{ \vtheta_0^0, \vdelta_1^0, \dots, \vdelta_n^0 \}$, each iteration (round $k+1$) updates the variables by iteratively solving the following single-variable minimization problem:
\begin{equation} \label{equ:3}
\begin{dcases}
\vtheta_0^{k+1} = \mathop{\arg\min}_{\vtheta} f(\vtheta, \vdelta_1^k ,\cdots, \vdelta_n^k) \\ 
\vdelta_i^{k+1} = \mathop{\arg\min}_{\vdelta} f(\cdots, \vdelta_{i-1}^k, \vdelta, \vdelta_{i+1}^k, \cdots), ~ \forall i
\end{dcases}
\end{equation}

The update for $\vtheta_0^*$ is trivial, while the update for $\vdelta$ is less straightforward due to the presence of the nuclear norm.
Fortunately, as shown in \citet{cai_2010_singular}, closed-form solutions for the coordinate descent method iterations in Problem (\ref{equ:2}) can be obtained using the Singular Value Thresholding (SVT) technique.
Recall that for a given matrix $\vdelta$ with the Singular Value Decomposition (SVD) $\vdelta = U \Sigma V^\top$, and a hyperparameter $\mu$, the SVT operator is defined as follows.
Let $\Sigma^+(\mu) := \text{diag}((\sigma_i - \mu)^+)$, where $(\cdot)^+$ denotes the positive part function. The SVT($\vdelta; \mu$) operator with hyperparameter $\mu$ is then defined as SVT($\vdelta; \mu$) := $U \Sigma^+ (\mu) V^\top$.
Using the SVT operator, the update for $\vdelta_i$ can be expressed as:
$\vdelta_i^{k+1} = $ SVT($\vtheta_i - \vtheta_0^{k+1}; \mu$).

Once the optimization problem is solved, we can obtain the approximate base model and a set of low-rank task vectors. Then, existing task-vector based approaches, such as Average Merging and TIES-Merging, can be applied to combine the task vectors and the base model. In this work, we directly adopt Average Merging as our post-calculation merging methods for simplicity, as as it demonstrated comparable performance to TIES-Merging in our preliminary experiments. The overall process is outlined in Algorithm \ref{algo:2}.

\section{Experiments} \label{sec:exp}

\begin{table*}[htbp]
\centering
\fontsize{8}{8} \selectfont
\def\arraystretch{1,2}
\begin{tabular}{c | cc | cc | ccc | ccc | c}
\toprule
\multirow{2}{*}{\textbf{Method}} & \multicolumn{2}{c|}{\textbf{DPSK \& Numina}} & \multicolumn{2}{c|}{\textbf{LM \& Math}} & \multicolumn{3}{c|}{\textbf{Math \& Code}} & \multicolumn{3}{c|}{\textbf{Checkpoints Merging}} & \multirow{2}{*}{\textbf{Avg.}} \\
\cline{2-3} \cline{4-5} \cline{6-8} \cline{9-11}
& {GSM8K} & {MATH} & {GSM8K} & {MATH} & MMLU & GLUE & MBPP & EasyLP & ComplexLP & NL4OPT \\
\hline
Baseline & 76.3 & 55.8 & 54.8 & 12.4 & 52.0 & 63.3 & 32.0 & 81.9 & 39.3 & 94.0 & 56.18 \\ 
Average & 75.0 & 45.8 & 58.8 & 12.6 & 52.8 & 61.7 & 28.0 & 75.9 & 40.3 & 91.6 & 54.25 \\ 
DARE & 81.0 &  54.2 & 14.9 & 3.7 & 52.7 & 59.1 & 27.6 & 80.7 & 35.1 & 95.1 & 50.41 \\ 
TIES & 80.8 & 51.6 & 58.5 & 11.8 & 53.1 & 59.3 & 26.8 & 82.4 & 42.7 & 94.8 & 56.18 \\
\hdashline
{\textsc{LoRE}} & 81.0 & 52.7 & 60.3 & 13.0 & 53.7 & 62.4 & 28.8 & 83.4 & 47.4 & 94.8 & \textbf{57.75} \\ 
\bottomrule
\end{tabular}
\caption{Evaluations on various benchmarks. LM and Math are Wizard-series models, namely WizardLM-13B and WizardMath-13B. Code is llama-2-13b-code-alpaca model. The score of baseline is the higher one of base models.}
\label{table:1}
\vspace{-3mm}
\end{table*}


\begin{table*}[t!]
\def\arraystretch{1,2}
\begin{minipage}[t]{.48\textwidth} 
\centering
\fontsize{8.5}{9}\selectfont
\begin{tabular}{c | c c c c}
\toprule
Datasets & $\mu=0$ & $\mu=0.01$ & $\mu=0.1$ & $\mu=1.0$ \\
\hline
GSM8K (\%) & 81.3 & 82.0 & 79.9 & 67.3 \\
MATH (\%) & 53.8 & 54.5 & 53.8 & 42.4 \\
\bottomrule
\end{tabular}
\caption{The ablation study for the hyperparameter $\mu$ (with $\lambda=1.0$) on DPSK \& Numina.}
\label{table:3}
\end{minipage}\hfill 
\begin{minipage}[t]{.48\textwidth} 
\centering
\def\arraystretch{1,2}
\fontsize{8.5}{9}\selectfont
\begin{tabular}{c | c c c}
\toprule
Datasets & $\lambda=0.5$ & $\lambda=1.0$ & $\lambda=1.5$ \\
\hline
GSM8K (\%) & 18.9 & 82.0 & 79.1 \\
MATH (\%) & 33.1 & 54.5 & 51.0 \\
\bottomrule
\end{tabular}
\caption{The ablation study for the hyperparameter $\lambda$ (with $\mu=0.01$) on DPSK \& Numina.}
\label{table:4}
\end{minipage}
\vspace{-3mm} 
\end{table*}

\paragraph{Baselines \& Settings}
We compare \textsc{LoRE-Merging} with following popular merging methods.
{\textbf{Average Merging}} \citep{choshen_2022_fusing}: This method computes the element-wise mean of all the individual models.
{\textbf{DARE}} \citep{yu_2024_language}: This approach randomly drops task-specific vectors and rescales the remaining vectors back to the base model. We set the hyperparameter for the random probability to $0.5$.
{\textbf{TIES-Merging}} \citep{yadav_2024_ties}: In this method, task-specific vectors are randomly dropped, and only the parameters aligned with the final agreed-upon sign are merged. For TIES-merging, we set the top-$k$ value to $20 \%$, and the hyperparameter $\lambda$ is fixed at $1$.
For \textsc{LoRE-Merging}, the rank $r$ is determined dynamically. 
For a given task vector $\vdelta \in \mathbf{R}^{m \times n}$, we set the rank $r = 0.2 \times \min \{ m, n \}$ to get a low-rank estimation.

\paragraph{Evaluation}
We first evaluate \textsc{LoRE-Merging} on diverse benchmarks, including GSM8K \citep{cobbe_2021_training}, MATH \citep{hendrycks2measuring} (math word problem), MMLU \citep{hendrycks2measuring}, GLUE\citep{wang2019glue} (commonsense reasoning) and MBPP\citep{austin2021program} (code task).
We evaluate DeepSeek-series models (NuminaMath-7B \citep{beeching_2024_numinamath} and DeepSeek-Math-7B-Base \citep{shao_2024_deepseekmath}) and LLaMA-series models (WizardLM-13B \citep{xu2023wizardlm}, WizardMath-13B \citep{luo2023wizardmath} and LLaMA-2-13B-Code model).
Additionally, we also evaluate on the advanced task, i.e. mathematical optimization modeling problems \citep{ramamonjison_2023_nl4opt, huang_2024_mamo, huang_2025_orlm}. This task aims to generate solvable mathematical models given an optimization problem in natural language. As the lack of public models on this task, we first fine-tuned Qwen-2.5-Coder-7B-Instruct model \citep{hui_2024_qwen2} with the dataset provided by \citet{huang_2025_orlm} and merge checkpoints in the training process. The evaluations are conducted on MAMO dataset \citep{huang_2024_mamo} which includes two subsets EasyLP and ComplexLP, and NL4OPT dataset \citep{ramamonjison_2023_nl4opt}.

\paragraph{Main Results}
As shown in Table \ref{table:1}, \textsc{LoRE-Merging} achieves superior performance across most metrics, as well as the highest overall score. For the math word problem evaluation, our method demonstrates consistently superior performance against baselines, except for the evaluations on MATH (DPSK \& Numina) and MBPP datasets. We think this is because of the significant performance gap between the base models, where DeepSeek-Math achieves only a score of 36.2 on the MATH dataset, while NuminaMath reaches 55.8. As indicated in \citet{yao2024determine}, a large performance gap can significantly impact the effectiveness of model merging. Another worthy-noting observation is that DARE demonstrates significantly poorer performance when merging WizardLM and WizardMath. This can likely be attributed to the substantial parameter divergence between these models, which results in the failure of calculating the task vector derived from the base model.
In contrast, our \textsc{LoRE-Merging} with the approximate base model and low-rank task vectors demonstrates superior robustness and effectiveness in solving math word problems.
For the evaluations on optimization modeling with checkpoints merging, we can see existing task-vector based merging methods consistently improve the performance because of the marginal gap between the checkpoints. Therefore, we believe that checkpoint merging can serve as a highly effective technique complementary to training methods, particularly our \textsc{LoRE-Merging} method. We also conduct a detailed analysis how our method enhance the modeling capacity on ComplexLP dataset. We found that the earlier checkpoint is more good at identifying the variables and parameters in the questions while the later one focuses on more complex components, such as formulating variables and the constraints. With the merging of task vectors, the merged model exhibits superior overall performance on the task.


\paragraph{Ablations}
We conduct a systematic empirical analysis of the selection of hyperparameters $\lambda$ and $\mu$, as presented in Table \ref{table:3} and Table \ref{table:4}. Our results show that the best performance is achieved with $\lambda = 1.0$ and $\mu = 0.01$. Notably, variations in the hyperparameters around these values do not significantly impact the final performance, indicating the robustness of \textsc{LoRE-Merging}.

\section{Conclusion}
In this paper, we propose a unified framework for merging models based on low-rank estimation, named \textsc{LoRE-Merging}.
We achieve it by formulating the merging problem as an optimization problem.
Extensive experiments demonstrate the efficacy and efficiency of our proposed methods.

\section*{Limitations}
Although we have demonstrated the effectiveness of our method on merging homogeneous models, we have not yet evaluated it on merging heterogeneous models which is a much more challenging task. Compared to existing task-vector based model merging methods, our method is the most suitable one that can be adapted to heterogeneous model merging, as we disentangle the base model and task vectors. We think how to expand \textsc{LoRE-Merging} to heterogeneous model merging should be a promising future direction.

\bibliography{custom.bib}

\begin{thebibliography}{37}
\providecommand{\natexlab}[1]{#1}

\bibitem[{Arpit et~al.(2022)Arpit, Wang, Zhou, and Xiong}]{arpit_2022_ensemble}
Devansh Arpit, Huan Wang, Yingbo Zhou, and Caiming Xiong. 2022.
\newblock Ensemble of averages: Improving model selection and boosting performance in domain generalization.
\newblock \emph{Advances in Neural Information Processing Systems}, 35:8265--8277.

\bibitem[{Austin et~al.(2021)Austin, Odena, Nye, Bosma, Michalewski, Dohan, Jiang, Cai, Terry, Le et~al.}]{austin2021program}
Jacob Austin, Augustus Odena, Maxwell Nye, Maarten Bosma, Henryk Michalewski, David Dohan, Ellen Jiang, Carrie Cai, Michael Terry, Quoc Le, and 1 others. 2021.
\newblock Program synthesis with large language models.
\newblock \emph{arXiv preprint arXiv:2108.07732}.

\bibitem[{Beeching et~al.(2024)Beeching, Huang, Jiang, Li, Lipkin, Qina, Rasul, Shen, Soletskyi, and Tunstall}]{beeching_2024_numinamath}
Edward Beeching, Shengyi~Costa Huang, Albert Jiang, Jia Li, Benjamin Lipkin, Zihan Qina, Kashif Rasul, Ziju Shen, Roman Soletskyi, and Lewis Tunstall. 2024.
\newblock Numinamath 7b tir.
\newblock \url{https://huggingface.co/AI-MO/NuminaMath-7B-TIR}.

\bibitem[{Bommasani et~al.(2021)Bommasani, Hudson, Adeli, Altman, Arora, von Arx, Bernstein, Bohg, Bosselut, Brunskill et~al.}]{bommasani_2021_opportunities}
Rishi Bommasani, Drew~A Hudson, Ehsan Adeli, Russ Altman, Simran Arora, Sydney von Arx, Michael~S Bernstein, Jeannette Bohg, Antoine Bosselut, Emma Brunskill, and 1 others. 2021.
\newblock On the opportunities and risks of foundation models.
\newblock \emph{arXiv preprint arXiv:2108.07258}.

\bibitem[{Cai et~al.(2008)Cai, Candes, and Shen}]{cai_2008_singular}
Jian-Feng Cai, Emmanuel~J. Candes, and Zuowei Shen. 2008.
\newblock \href {https://arxiv.org/abs/0810.3286} {A singular value thresholding algorithm for matrix completion}.
\newblock \emph{Preprint}, arXiv:0810.3286.

\bibitem[{Cai et~al.(2010)Cai, Cand{\`e}s, and Shen}]{cai_2010_singular}
Jian-Feng Cai, Emmanuel~J Cand{\`e}s, and Zuowei Shen. 2010.
\newblock A singular value thresholding algorithm for matrix completion.
\newblock \emph{SIAM Journal on optimization}, 20(4):1956--1982.

\bibitem[{Cha et~al.(2021)Cha, Chun, Lee, Cho, Park, Lee, and Park}]{cha_2021_swad}
Junbum Cha, Sanghyuk Chun, Kyungjae Lee, Han-Cheol Cho, Seunghyun Park, Yunsung Lee, and Sungrae Park. 2021.
\newblock Swad: Domain generalization by seeking flat minima.
\newblock \emph{Advances in Neural Information Processing Systems}, 34:22405--22418.

\bibitem[{Choshen et~al.(2022)Choshen, Venezian, Slonim, and Katz}]{choshen_2022_fusing}
Leshem Choshen, Elad Venezian, Noam Slonim, and Yoav Katz. 2022.
\newblock Fusing finetuned models for better pretraining.
\newblock \emph{arXiv preprint arXiv:2204.03044}.

\bibitem[{Cobbe et~al.(2021)Cobbe, Kosaraju, Bavarian, Chen, Jun, Kaiser, Plappert, Tworek, Hilton, Nakano, Hesse, and Schulman}]{cobbe_2021_training}
Karl Cobbe, Vineet Kosaraju, Mohammad Bavarian, Mark Chen, Heewoo Jun, Lukasz Kaiser, Matthias Plappert, Jerry Tworek, Jacob Hilton, Reiichiro Nakano, Christopher Hesse, and John Schulman. 2021.
\newblock Training verifiers to solve math word problems.
\newblock \emph{arXiv preprint arXiv:2110.14168}.

\bibitem[{Don-Yehiya et~al.(2022)Don-Yehiya, Venezian, Raffel, Slonim, Katz, and Choshen}]{don_2022_cold}
Shachar Don-Yehiya, Elad Venezian, Colin Raffel, Noam Slonim, Yoav Katz, and Leshem Choshen. 2022.
\newblock Cold fusion: Collaborative descent for distributed multitask finetuning.
\newblock \emph{arXiv preprint arXiv:2212.01378}.

\bibitem[{Garipov et~al.(2018)Garipov, Izmailov, Podoprikhin, Vetrov, and Wilson}]{garipov_2018_loss}
Timur Garipov, Pavel Izmailov, Dmitrii Podoprikhin, Dmitry~P Vetrov, and Andrew~G Wilson. 2018.
\newblock Loss surfaces, mode connectivity, and fast ensembling of dnns.
\newblock \emph{Advances in neural information processing systems}, 31.

\bibitem[{Gupta et~al.(2020)Gupta, Serrano, and DeCoste}]{gupta_2020_stochastic}
Vipul Gupta, Santiago~Akle Serrano, and Dennis DeCoste. 2020.
\newblock Stochastic weight averaging in parallel: Large-batch training that generalizes well.
\newblock \emph{arXiv preprint arXiv:2001.02312}.

\bibitem[{Hendrycks et~al.()Hendrycks, Burns, Kadavath, Arora, Basart, Tang, Song, and Steinhardt}]{hendrycks2measuring}
Dan Hendrycks, Collin Burns, Saurav Kadavath, Akul Arora, Steven Basart, Eric Tang, Dawn Song, and Jacob Steinhardt.
\newblock Measuring mathematical problem solving with the math dataset.
\newblock In \emph{Thirty-fifth Conference on Neural Information Processing Systems Datasets and Benchmarks Track (Round 2)}.

\bibitem[{Huang et~al.(2025)Huang, Tang, Hu, Jiang, Zheng, Ge, Wang, and Wang}]{huang_2025_orlm}
Chenyu Huang, Zhengyang Tang, Shixi Hu, Ruoqing Jiang, Xin Zheng, Dongdong Ge, Benyou Wang, and Zizhuo Wang. 2025.
\newblock \href {https://arxiv.org/abs/2405.17743} {Orlm: A customizable framework in training large models for automated optimization modeling}.
\newblock \emph{Preprint}, arXiv:2405.17743.

\bibitem[{Huang et~al.(2024)Huang, Shen, Hu, Gao, and Wang}]{huang_2024_mamo}
Xuhan Huang, Qingning Shen, Yan Hu, Anningzhe Gao, and Benyou Wang. 2024.
\newblock \href {https://arxiv.org/abs/2405.13144} {Mamo: a mathematical modeling benchmark with solvers}.
\newblock \emph{Preprint}, arXiv:2405.13144.

\bibitem[{Hui et~al.(2024)Hui, Yang, Cui, Yang, Liu, Zhang, Liu, Zhang, Yu, Dang et~al.}]{hui_2024_qwen2}
Binyuan Hui, Jian Yang, Zeyu Cui, Jiaxi Yang, Dayiheng Liu, Lei Zhang, Tianyu Liu, Jiajun Zhang, Bowen Yu, Kai Dang, and 1 others. 2024.
\newblock Qwen2. 5-coder technical report.
\newblock \emph{arXiv preprint arXiv:2409.12186}.

\bibitem[{Ilharco et~al.(2022)Ilharco, Ribeiro, Wortsman, Gururangan, Schmidt, Hajishirzi, and Farhadi}]{ilharco_2022_editing}
Gabriel Ilharco, Marco~Tulio Ribeiro, Mitchell Wortsman, Suchin Gururangan, Ludwig Schmidt, Hannaneh Hajishirzi, and Ali Farhadi. 2022.
\newblock Editing models with task arithmetic.
\newblock \emph{arXiv preprint arXiv:2212.04089}.

\bibitem[{Jin et~al.(2022)Jin, Ren, Preotiuc-Pietro, and Cheng}]{jin_2022_dataless}
Xisen Jin, Xiang Ren, Daniel Preotiuc-Pietro, and Pengxiang Cheng. 2022.
\newblock Dataless knowledge fusion by merging weights of language models.
\newblock \emph{arXiv preprint arXiv:2212.09849}.

\bibitem[{Li et~al.(2022)Li, Gururangan, Dettmers, Lewis, Althoff, Smith, and Zettlemoyer}]{li_2022_branch}
Margaret Li, Suchin Gururangan, Tim Dettmers, Mike Lewis, Tim Althoff, Noah~A Smith, and Luke Zettlemoyer. 2022.
\newblock Branch-train-merge: Embarrassingly parallel training of expert language models.
\newblock \emph{arXiv preprint arXiv:2208.03306}.

\bibitem[{Liu et~al.(2024)Liu, Wang, Wang, Chen, Li, Tu, Chu, Li, and Sui}]{liu_2024_checkpoint}
Deyuan Liu, Zecheng Wang, Bingning Wang, Weipeng Chen, Chunshan Li, Zhiying Tu, Dianhui Chu, Bo~Li, and Dianbo Sui. 2024.
\newblock \href {https://arxiv.org/abs/2403.19390} {Checkpoint merging via bayesian optimization in llm pretraining}.
\newblock \emph{Preprint}, arXiv:2403.19390.

\bibitem[{Lu et~al.(2024)Lu, Fan, Wei, Qu, Chen, and Cheng}]{lu_2024_twin}
Zhenyi Lu, Chenghao Fan, Wei Wei, Xiaoye Qu, Dangyang Chen, and Yu~Cheng. 2024.
\newblock \href {https://proceedings.neurips.cc/paper_files/paper/2024/file/8fcd17eb91bae20d9826786d7d6be799-Paper-Conference.pdf} {Twin-merging: Dynamic integration of modular expertise in model merging}.
\newblock In \emph{Advances in Neural Information Processing Systems}, volume~37, pages 78905--78935. Curran Associates, Inc.

\bibitem[{Luo et~al.(2023)Luo, Sun, Xu, Zhao, Lou, Tao, Geng, Lin, Chen, and Zhang}]{luo2023wizardmath}
Haipeng Luo, Qingfeng Sun, Can Xu, Pu~Zhao, Jianguang Lou, Chongyang Tao, Xiubo Geng, Qingwei Lin, Shifeng Chen, and Dongmei Zhang. 2023.
\newblock Wizardmath: Empowering mathematical reasoning for large language models via reinforced evol-instruct.
\newblock \emph{arXiv preprint arXiv:2308.09583}.

\bibitem[{Matena and Raffel(2022)}]{matena_2022_merging}
Michael~S Matena and Colin~A Raffel. 2022.
\newblock Merging models with fisher-weighted averaging.
\newblock \emph{Advances in Neural Information Processing Systems}, 35:17703--17716.

\bibitem[{Ramamonjison et~al.(2023)Ramamonjison, Yu, Li, Li, Carenini, Ghaddar, He, Mostajabdaveh, Banitalebi-Dehkordi, Zhou, and Zhang}]{ramamonjison_2023_nl4opt}
Rindranirina Ramamonjison, Timothy~T. Yu, Raymond Li, Haley Li, Giuseppe Carenini, Bissan Ghaddar, Shiqi He, Mahdi Mostajabdaveh, Amin Banitalebi-Dehkordi, Zirui Zhou, and Yong Zhang. 2023.
\newblock \href {https://arxiv.org/abs/2303.08233} {Nl4opt competition: Formulating optimization problems based on their natural language descriptions}.
\newblock \emph{Preprint}, arXiv:2303.08233.

\bibitem[{Ram{\'e} et~al.(2023)Ram{\'e}, Ahuja, Zhang, Cord, Bottou, and Lopez-Paz}]{rame_2023_model}
Alexandre Ram{\'e}, Kartik Ahuja, Jianyu Zhang, Matthieu Cord, L{\'e}on Bottou, and David Lopez-Paz. 2023.
\newblock Model ratatouille: Recycling diverse models for out-of-distribution generalization.
\newblock In \emph{International Conference on Machine Learning}, pages 28656--28679. PMLR.

\bibitem[{Shao et~al.(2024)Shao, Wang, Zhu, Xu, Song, Bi, Zhang, Zhang, Li, Wu, and Guo}]{shao_2024_deepseekmath}
Zhihong Shao, Peiyi Wang, Qihao Zhu, Runxin Xu, Junxiao Song, Xiao Bi, Haowei Zhang, Mingchuan Zhang, Y.~K. Li, Y.~Wu, and Daya Guo. 2024.
\newblock \href {https://arxiv.org/abs/2402.03300} {Deepseekmath: Pushing the limits of mathematical reasoning in open language models}.
\newblock \emph{Preprint}, arXiv:2402.03300.

\bibitem[{Wang et~al.(2019)Wang, Singh, Michael, Hill, Levy, and Bowman}]{wang2019glue}
Alex Wang, Amanpreet Singh, Julian Michael, Felix Hill, Omer Levy, and Samuel~R. Bowman. 2019.
\newblock {GLUE}: A multi-task benchmark and analysis platform for natural language understanding.
\newblock In the Proceedings of ICLR.

\bibitem[{Wortsman et~al.(2022)Wortsman, Ilharco, Gadre, Roelofs, Gontijo-Lopes, Morcos, Namkoong, Farhadi, Carmon, Kornblith et~al.}]{wortsman_2022_model}
Mitchell Wortsman, Gabriel Ilharco, Samir~Ya Gadre, Rebecca Roelofs, Raphael Gontijo-Lopes, Ari~S Morcos, Hongseok Namkoong, Ali Farhadi, Yair Carmon, Simon Kornblith, and 1 others. 2022.
\newblock Model soups: averaging weights of multiple fine-tuned models improves accuracy without increasing inference time.
\newblock In \emph{International conference on machine learning}, pages 23965--23998. PMLR.

\bibitem[{Wright(2015)}]{wright_2015_coordinate}
Stephen~J. Wright. 2015.
\newblock \href {https://arxiv.org/abs/1502.04759} {Coordinate descent algorithms}.
\newblock \emph{Preprint}, arXiv:1502.04759.

\bibitem[{Xu et~al.(2023)Xu, Sun, Zheng, Geng, Zhao, Feng, Tao, and Jiang}]{xu2023wizardlm}
Can Xu, Qingfeng Sun, Kai Zheng, Xiubo Geng, Pu~Zhao, Jiazhan Feng, Chongyang Tao, and Daxin Jiang. 2023.
\newblock Wizardlm: Empowering large language models to follow complex instructions.
\newblock \emph{arXiv preprint arXiv:2304.12244}.

\bibitem[{Yadav and Bansal(2022)}]{yadav_2022_exclusive}
Prateek Yadav and Mohit Bansal. 2022.
\newblock Exclusive supermask subnetwork training for continual learning.
\newblock \emph{arXiv preprint arXiv:2210.10209}.

\bibitem[{Yadav et~al.(2023)Yadav, Sun, Ding, Li, Zhang, Tan, Ma, Bhatia, Nallapati, Ramanathan et~al.}]{yadav_2023_exploring}
Prateek Yadav, Qing Sun, Hantian Ding, Xiaopeng Li, Dejiao Zhang, Ming Tan, Xiaofei Ma, Parminder Bhatia, Ramesh Nallapati, Murali~Krishna Ramanathan, and 1 others. 2023.
\newblock Exploring continual learning for code generation models.
\newblock \emph{arXiv preprint arXiv:2307.02435}.

\bibitem[{Yadav et~al.(2024)Yadav, Tam, Choshen, Raffel, and Bansal}]{yadav_2024_ties}
Prateek Yadav, Derek Tam, Leshem Choshen, Colin~A Raffel, and Mohit Bansal. 2024.
\newblock Ties-merging: Resolving interference when merging models.
\newblock \emph{Advances in Neural Information Processing Systems}, 36.

\bibitem[{Yao et~al.(2024)Yao, Wu, Liu, Luo, Han, Liu, Guo, and Song}]{yao2024determine}
Yuxuan Yao, Han Wu, Mingyang Liu, Sichun Luo, Xiongwei Han, Jie Liu, Zhijiang Guo, and Linqi Song. 2024.
\newblock Determine-then-ensemble: Necessity of top-k union for large language model ensembling.
\newblock \emph{arXiv preprint arXiv:2410.03777}.

\bibitem[{Yu et~al.(2024{\natexlab{a}})Yu, Yu, Yu, Huang, and Li}]{yu2024extend}
Le~Yu, Bowen Yu, Haiyang Yu, Fei Huang, and Yongbin Li. 2024{\natexlab{a}}.
\newblock Extend model merging from fine-tuned to pre-trained large language models via weight disentanglement.
\newblock \emph{arXiv preprint arXiv:2408.03092}.

\bibitem[{Yu et~al.(2024{\natexlab{b}})Yu, Yu, Yu, Huang, and Li}]{yu_2024_language}
Le~Yu, Bowen Yu, Haiyang Yu, Fei Huang, and Yongbin Li. 2024{\natexlab{b}}.
\newblock Language models are super mario: Absorbing abilities from homologous models as a free lunch.
\newblock In \emph{Forty-first International Conference on Machine Learning}.

\bibitem[{Zhuang et~al.(2020)Zhuang, Qi, Duan, Xi, Zhu, Zhu, Xiong, and He}]{zhuang_2020_comprehensive}
Fuzhen Zhuang, Zhiyuan Qi, Keyu Duan, Dongbo Xi, Yongchun Zhu, Hengshu Zhu, Hui Xiong, and Qing He. 2020.
\newblock A comprehensive survey on transfer learning.
\newblock \emph{Proceedings of the IEEE}, 109(1):43--76.

\end{thebibliography}

\clearpage
\appendix

\section{Appendix}
\begin{table*}[!t]
    \centering
    \def\arraystretch{1,2}
    \begin{tabular}{ccccc}
    \toprule
      Method & Average & TIES-Merging & Twin-Merging & LoRE-Merging \\
      \hline
      Acc. on GSM8K & 75.0 & 80.8 & 79.9 & 81.0\\
      Runtime & 4.2s & 5min 29s & 17min 44s & 12min 24s\\
      \bottomrule
    \end{tabular}
    \caption{Time cost comparisons among different merging methods.}
    \label{tab:speed}
\end{table*}

\subsection{Speed and Computational Cost}

While standard SVD exhibits computational inefficiency for extremely large matrices comprising billions of elements, its application to LLM presents a substantially different computational profile. Despite LLMs containing billions of parameters in aggregate, SVD operations are performed on individual parameter matrices, each typically comprising only millions of entries. For instance, in the Qwen2.5-72B architecture, the largest matrix requiring decomposition is dimensioned at $8192 \times 28564$, while for Qwen2.5-7B, the corresponding matrix has dimensions of $3854 \times 18944$. Thus, the substantial parameter differential between LLM scales does not translate to proportionally expanded matrix dimensions. In our implementation, merging operations for 7B-scale models require approximately 5 minutes using Ties-Merging, while LoRE-Merging necessitates approximately 12 minutes. However, compared to another SVD-based mering method, like Twin-Merging \citep{lu_2024_twin}, our method exhibit superior performance on efficiency.

\subsection{Task Vector Rank Validation}

In this subsection, we validate the low-rank properties underlying the low-rank assumption. Specifically, we focus on the checkpoint merging problem and compute the rank of the task vectors. As previously discussed, we set the rank $r$ as $r = 0.2 \times \min \{ m, n \}$ for any given task vector $\vdelta$.

The distribution of the largest 100 singular values for Layer $1$ is presented in Figure \ref{fig:1}. Our experimental results reveal that $\sigma_r \leq 0.05 \times \sigma_1$, indicating that the singular values set to $0$ in low-rank estimation are significantly smaller than the largest singular value across all linear layers. This finding supports the validity of adopting a low-rank approximation for task vectors, as it reflects the inherent structure of the data.

\begin{algorithm}[htbp]
\caption{Implicit low-rank merging method}
\label{algo:2}
Input: fine-tuned models $\{ \vtheta_i \}_{i=1}^n$, parameter dimension $d$, and hyperparameter $\lambda, \mu$.

Output: merged model $\vtheta^*$.

\begin{algorithmic}
\STATE $\rhd$ Step 1: Coordinate descent method to solve problem (\ref{equ:2}).
\STATE Set $\vdelta_i = 0$ for $i = 1, 2, \dots, n$.
\WHILE{iteration NOT converges}
\STATE $\vtheta_0 = \frac{1}{n} \sum_{i=1}^n (\vtheta_i - \vdelta_i)$
\FOR{$i=1, \dots, n$}
\STATE $\vdelta_i = \mathrm{SVT} (\vtheta_i - \vtheta_0; \mu)$;
\ENDFOR 
\ENDWHILE
\STATE
\STATE $\rhd$ Step 2 (Optional 1): Direct sum.
\STATE $\vtau = \sum_{i=1}^n \vdelta_i$.
\STATE
\STATE $\rhd$ Step 2 (Optional 2): TIES selection \citep{yadav_2024_ties}.
\STATE $\vgamma = sgn(\sum_{i=1}^n \vdelta_i)$.
\FOR{$p = 1, 2, \dots, d$}
\STATE $\mathcal{A}^p = \{i:  \vgamma_i^p = \vgamma^p \}$
\STATE $\vtau^p = \frac{1}{|\mathcal{A}^p|} \sum_{i \in \mathcal{A}^p} \vtau^p$
\ENDFOR
\STATE
\STATE $\rhd$ Step 3: Obtain merged checkpoint.
\STATE $\vtheta^* = \vtheta_0 + \lambda \vtau$.
\RETURN $\vtheta^*$
\end{algorithmic}
\end{algorithm}

\end{document}